%% file: main.tex
\documentclass[10pt,twocolumn,letterpaper]{article}

\usepackage{iccv}
\usepackage{times}
\usepackage{epsfig}
\usepackage{graphicx}
\usepackage{amsmath}
\usepackage{amssymb}
\usepackage{multirow}
\usepackage{makecell}

\usepackage[pagebackref=true,breaklinks=true,letterpaper=true,colorlinks,bookmarks=false]{hyperref}

\iccvfinalcopy 

\def\iccvPaperID{3} 

\ificcvfinal\fi
\begin{document}

\title{Do Deep Neural Networks Learn Facial Action Units \\When Doing Expression Recognition?}

\author{Pooya Khorrami \hspace{2cm} Tom Le Paine \hspace{2cm} Thomas S. Huang\\
Beckman Institute for Advanced Science and Technology\\
University of Illinois at Urbana-Champaign\\
\texttt{\{pkhorra2,paine1,t-huang1\}@illinois.edu} 
}


\maketitle
\thispagestyle{empty}

\begin{abstract}
Despite being the appearance-based classifier of choice in recent years, relatively few works have examined how much convolutional neural networks (CNNs) can improve performance on accepted expression recognition benchmarks and, more importantly, examine what it is they actually learn. In this work, not only do we show that CNNs can achieve strong performance, but we also introduce an approach to decipher which portions of the face influence the CNN's predictions. First, we train a zero-bias CNN on facial expression data and achieve, to our knowledge, state-of-the-art performance on two expression recognition benchmarks: the extended Cohn-Kanade (CK+) dataset and the Toronto Face Dataset (TFD). We then qualitatively analyze the network by visualizing the spatial patterns that maximally excite different neurons in the convolutional layers and show how they resemble Facial Action Units (FAUs). Finally, we use the FAU labels provided in the CK+ dataset to verify that the FAUs observed in our filter visualizations indeed align with the subject's facial movements.
\end{abstract}


\section{Introduction}
\label{sec:intro}
\input{./Sections/intro.tex}

\section{Related Work}
\label{sec:related_work}
\input{./Sections/related_work.tex}

\section{Our Approach}
\label{sec:method}
\input{./Sections/method.tex}

\section{Experiments and Analysis}
\label{sec:experiments}
\input{./Sections/experiments.tex}

\section{Conclusions}
\label{sec:conclusions}
\input{./Sections/conclusions.tex}

\section*{Acknowledgments}
This work was supported in part by MIT Lincoln Laboratory. 
The Tesla K40 GPU used for this research was donated by the NVIDIA Corporation. The authors would also like to thank Dr. Kevin Brady, Dr. Charlie Dagli, Professor Yun Fu, and Professor Usman Tariq for their insightful comments and suggestions with regards to this work.

{\small
\bibliographystyle{ieee}
\bibliography{egbib}
}

\end{document}

%% file: Sections/intro.tex
Facial expressions provide a natural and compact way for humans to convey their emotional state to another party. Therefore, designing accurate facial expression recognition algorithms is crucial to the development of interactive computer systems in artificial intelligence. Extensive work in this area has found that only a small number of regions change as a human changes their expression and are located around the subject's eyes, nose and mouth. In \cite{ekman1977facial}, Paul Ekman proposed the Facial Action Coding System (FACS) which enumerated these regions and described how every facial expression can be described as the combination of multiple action units (AUs), each corresponding to a particular muscle group in the face. However, having a computer accurately learn the parts of the face that convey emotion has proven to be a non-trivial task. 

\begin{figure}[t!]
  \centering
  \centerline{\includegraphics[width=8cm, height=8cm]{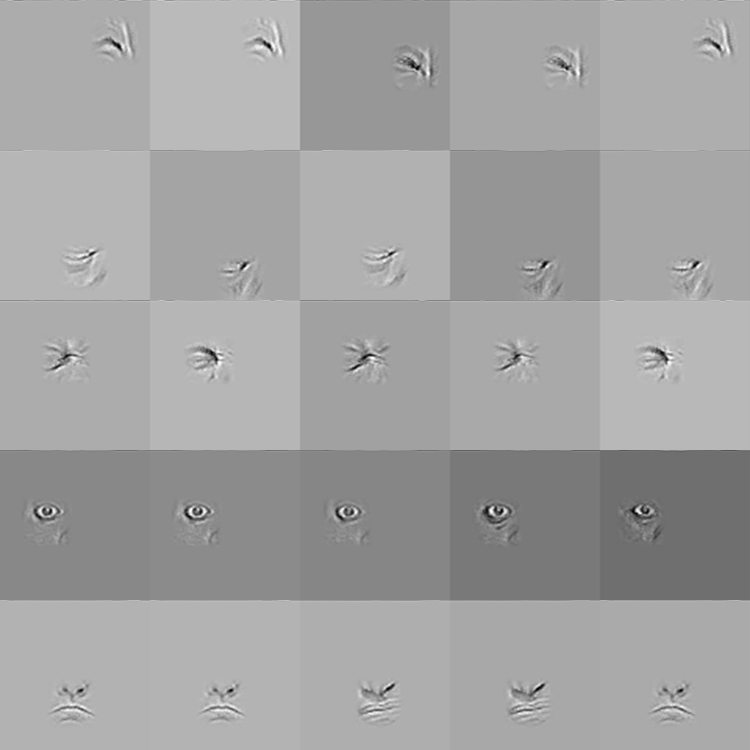}}
  \vspace{0.2cm}
  \caption{Visualization of facial regions that activate five selected filters in the 3rd convolutional layer of a network trained on the Extended Cohn-Kanade (CK+) dataset. Each row corresponds to one filter in the conv3 layer and we display the spatial patterns from the top 5 images.}
\label{fig:zeiler_plots_ck_front}
\end{figure}

Previous work in facial expression recognition can be split into two broad categories: AU-based/rule-based methods and appearance-based methods. AU-based methods \cite{tian2000recognizing, tong2007facial} would detect the presence of individual AUs explicitly and then classify a person's emotion based on the combinations originally proposed by Friesen and Ekman in \cite{friesen1983emfacs}. Unfortunately, each AU detector required careful hand-engineering to ensure good performance. On the other hand, appearance-based methods \cite{bartlett2005recognizing, bartlett2006fully, whitehill2006haar, zhao2007dynamic} modeled a person's expression from their general facial shape and texture.

In the last few years, many well-established problems in computer vision have greatly benefited from the rise of convolutional neural networks (CNNs) as an appearance-based classifier. Tasks such as object recognition \cite{krizhevsky2012imagenet}, object detection \cite{girshick2014rich}, and face recognition \cite{taigman2014deepface} have seen huge boosts in performance on several accepted benchmarks. Unfortunately, other tasks such as facial expression recognition have not experienced performance gains of the same magnitude. Little work has been done to see how much deep CNNs can help on accepted expression recognition benchmarks.


In this paper, we seek the answer to the following questions: Can CNNs improve performance on emotion recognition datasets/baselines and what do they learn? We propose to do this by training a CNN on established facial expression datasets and then analyzing what they learn by visualizing the individual filters in the network. In this work, we apply the visualization techniques proposed by Zeiler and Fergus \cite{zeiler2014visualizing} and Springenberg et al. \cite{springenberg2014striving} where individual neurons in the network are excited and their corresponding spatial patterns are displayed in pixel space using a deconvolutional network. When visualizing these discriminative spatial patterns, we find that many of the filters are excited by regions in the face that corresponded to Facial Action Units (FAUs). A subset of these spatial patterns is shown in Figure \ref{fig:zeiler_plots_ck_front}.



Thus, the main contributions of this paper are as follows:
\begin{enumerate}
\item We show that CNNs trained for the emotion recognition task learn features that correspond strongly with the FAUs proposed by Ekman \cite{ekman1977facial}. We demonstrate this result by first visualizing the spatial patterns that maximally excite different filters in the convolutional layers of our networks, and then using the ground truth FAU labels to verify that the FAUs observed in the filter visualizations align with the subject's facial movements.

\item We also show that our CNN model, based on works originally proposed by \cite{memisevic2014zero, paine2014analysis}, can achieve, to our knowledge, state-of-the-art performance on the extended Cohn-Kanade (CK+) dataset and the Toronto Face Dataset (TFD). 
\end{enumerate}

%% file: Sections/related_work.tex
\begin{figure*}[t!]
\begin{minipage}[b]{\linewidth}
  \centering
 \centerline{\includegraphics[trim = 0mm 40mm 0mm 60mm, clip, width=16cm, height=5cm, keepaspectratio]{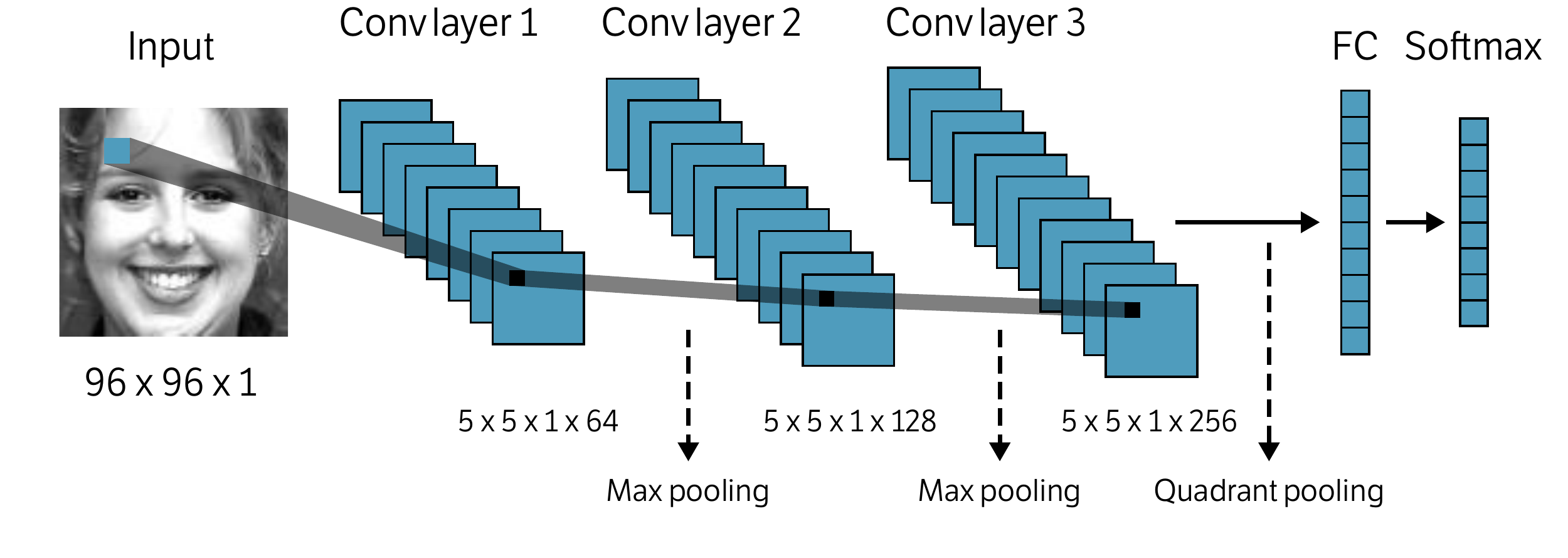}}  
\end{minipage}
\caption{Network Architecture - Our network consists of three convolutional layers containing 64, 128, and 256 filters, respectively, each of size 5x5 followed by ReLU (Rectified Linear Unit) activation functions. We add 2x2 max pooling layers after the first two convolutional layers and quadrant pooling after the third. The three convolutional layers are followed by a fully-connected layer containing 300 hidden units and a softmax layer.}
\label{fig:network_architecture}
\end{figure*} 

In most facial expression recognition systems, the main machinery matches quite nicely with the traditional machine learning pipeline. More specifically, a face image is passed to a classifier that tries to categorize it as one of several (typically 7) expression classes: 1. anger, 2. disgust, 3. fear, 4. neutral, 5. happy, 6. sad, and 7. surprise. In most cases, prior to being passed to the classifier, the face image is pre-processed and given to a feature extractor. Up until rather recently, most appearance-based expression recognition techniques relied on hand-crafted features, specifically Gabor wavelets \cite{bartlett2005recognizing, bartlett2006fully}, Haar features \cite{whitehill2006haar} and LBP features \cite{zhao2007dynamic}, in order to make representations of different expression classes more discriminative. 

For some time, systems based on hand-crafted features were able to achieve impressive results on accepted expression recognition benchmarks such as the Japanese Female Facial Expression (JAFFE) database \cite{lyons1999automatic}, the extended Cohn-Kanade (CK+) dataset \cite{lucey2010extended}, and the Multi-PIE dataset \cite{gross2010multi}. However, the recent success of deep neural networks has caused many researchers to explore feature representations that are learned from data. Not surprisingly, almost all of the methods used some form of unsupervised pre-training/learning to initialize their models. We hypothesize this may be because the scarcity of labeled data prevented the authors from training a completely supervised model that did not experience heavy overfitting. 

In \cite{liu2014facial}, the authors trained a multi-layer boosted deep belief network (BDBN) and achieved state-of-the-art accuracy on the CK+ and JAFFE datasets. Meanwhile in \cite{rifai2012disentangling}, the authors used a convolutional contractive auto-encoder (CAE) as their underlying unsupervised model. They then performed a semi-supervised encoding function called Contractive Discriminant Analysis (CDA) to separate discriminative expression features from the unsupervised representation. 

A few works based on unsupervised deep learning have also tried to analyze the relationship between FAUs and the learned feature representations. In \cite{liu2013aware, liu2015inspired}, the authors learned a patch-based filter bank using K-means as their low-level feature. These features were then used to select receptive fields corresponding to specific FAU receptive fields which were subsequently passed to multi-layer restricted Boltzmann machines (RBMs) for classification. The FAU receptive fields were selected using a mutual information criterion between the image feature and the expression label. An earlier work by Susskind et al. \cite{susskind2008generating}, showed that the first layer features a deep belief network trained to generate facial expression images appeared to learn filters that were sensitive to face parts. We conduct a similar analysis except we use a CNN as our underlying model and we visualize the spatial patterns that excite higher-level neurons in the network.

To the authors' knowledge, the only works that previously applied CNNs to expression data were that of Kahou et al. \cite{kahou2013combining, kahou2015emonets} and Jung et al. \cite{jung2015deep}. 
In \cite{kahou2013combining, kahou2015emonets}, the authors developed a system for doing audio/visual emotion recognition for the Emotion Recognition in the Wild Challenge (EmotiW) \cite{dhall2013emotion, dhall2014emotion} while in \cite{jung2015deep}, the authors trained a network that incorporated both appearance and geometric features when doing recognition. However, one key point is that these works dealt with emotion recognition of video / image sequence data and therefore, actively incorporated temporal data when computing their predictions. 

In contrast, our work deals with emotion recognition from a single image, and will focus on analyzing the features learned by the network. Thus, not only will we demonstrate the effectiveness of CNNs on existing emotion classification baselines but we will also qualitatively show that the network is able to learn patterns in the face images that correspond to Facial Action Units (FAUs).


%% file: Sections/method.tex
\subsection{Network Architecture}
For all of the experiments we present in this paper, we use a classic feed-forward convolutional neural network.  The networks we use, shown visually in Figure \ref{fig:network_architecture} consist of three convolutional layers with 64, 128, and 256 filters, respectively, and with filter sizes of 5x5 followed by ReLU (Rectified Linear Unit) activation functions. Max pooling layers are placed after the first two convolutional layers while quadrant pooling \cite{coates2011analysis} is applied after the third. The quadrant pooling layer is then followed by a full-connected layer with 300 hidden units and, finally, a softmax layer for classification. The softmax layer contains anywhere between 6-8 outputs corresponding to the number of expressions present in the training set.

One modification that we introduce to the classical configuration is that we ignore the biases of the convolutional layers. This idea was introduced first by Memisevic et al. in \cite{memisevic2014zero} for fully-connected networks and later extended by Paine et al. in \cite{paine2014analysis} to convolutional layers. 
In our experiments, we found that ignoring the bias allowed our network to train very quickly while simultaneously reducing the number of parameters to learn.
 
\subsection{Network Training}

When training our network, we train from scratch using stochastic gradient descent with a batch size of 64, momentum set to 0.9, and a weight decay parameter of 1e-5. We use a constant learning rate of 0.01 and do not use any form of annealing. The parameters of each layer are randomly initialized by drawing from a Gaussian distribution with zero mean and standard deviation $ \sigma=\frac{k}{ N_{\text{FAN\_IN}}}$ where $N_{\text{FAN\_IN}}$ is the number of input connections to each layer and k is drawn uniformly from the range: $\left[0.2, 1.2\right]$.

We also use dropout and various forms of data augmentation to regularize our network and combat overfitting. We apply dropout to the fully-connected layer with a probability of 0.5 (i.e. each neuron's output is set to zero with probability 0.5). For data augmentation, we apply a random transformation to each input image  consisting of: translations, horizontal flips, rotations, scaling, and pixel intensity augmentation. All of our models were trained using the anna software library \footnote{\url{https://github.com/ifp-uiuc/anna}}. 

%

%% file: Sections/experiments.tex
We use two facial expression datasets in our experiments: the extended Cohn-Kanade database (CK+) \cite{lucey2010extended} and the Toronto Face Dataset (TFD) \cite{susskind2010toronto}. The CK+ database contains 327 image sequences, each of which is assigned one of 7 expression labels: anger, contempt, disgust, fear, happy, sad, and surprise. For fair comparison, we follow the protocol used by previous works \cite{liu2013aware, liu2014facial}, and use the first frame of each sequence as a neutral frame in addition to the last three expressive frames to form our dataset. This leads to a total of 1308 images and 8 classes total. We then split the frames into 10 subject independent subsets in the manner presented by \cite{liu2013aware} and perform 10-fold cross-validation.

TFD is an amalgamation of several facial expression datasets. It contains 4178 images annotated with one of 7 expression labels: anger, disgust, fear, happy, neutral, sad, and surprise. The labeled samples are divided into 5 folds, each containing a train, validation, and test set. We train all of our models using just the training set of each fold, pick the best performing model using each split's validation set, then we evaluate on each split's test set and average the results over all 5 folds.

In both datasets, the images are grayscale and are of size 96x96 pixels. In the case of TFD, the faces have already been detected and normalized such that all of the subjects' eyes are the same distance apart and have the same vertical coordinates. Meanwhile for the CK+ dataset, we simply detect the face in the 640x480 image and resize it to 96x96. The only other pre-processing we employ is patch-wise mean subtraction and scaling to unit variance.

\subsection{Performance on Toronto Face Database (TFD)}
 
First, we analyze the discriminative ability of the CNN by assessing its performance on the TFD dataset.  
Table \ref{tab:tfd} shows the recognition accuracy obtained when training a zero-bias CNN from a random initialization with no other regularization as well as CNNs that have dropout (D), data augmentation (A) or both (AD). We also include recognition accuracies from previous methods. From the results in Table \ref{tab:tfd}, there are two main observations: (i) not surprisingly, regularization significantly boosts performance  (ii) data augmentation improves performance over the regular CNN more than dropout ($9.4\%$ vs. $2.8\%$). Furthermore, when both dropout and data augmentation are used, our model is able to exceed the previous state-of-the-art performance on TFD by $3.6\%$.

\begin{table}[t!]
\caption{Recognition accuracy on the Toronto Face Dataset (TFD) - 7 classes - A: Data Augmentation, D: Dropout}
\begin{center}
    \begin{tabular}{ | l | c  |}
    \hline
    \textbf{Method} & \textbf{Accuracy} \\ \hline
    Gabor+PCA \cite{dailey2002empath} & 80.2\% \\ \hline
    Deep mPoT \cite{ranzato2011deep} & 82.4\% \\ \hline
    CDA \cite{rifai2012disentangling} & 85.0\% \\ \hline
    \hline
    Zero-bias CNN & 79.0\% $\pm$ 1.1\%\\ \hline
    Zero-bias CNN+D & 81.8\% $\pm$ 2.1\% \\ \hline
    Zero-bias CNN+A & 88.4\% $\pm$ 1.7\% \\ \hline
    \textbf{Zero-bias CNN+AD} & \textbf{88.6\% $\pm$ 1.5\%} \\  \hline
    \end{tabular}
\end{center}
\label{tab:tfd}
\end{table}

\subsection{Performance on the Extended Cohn-Kanade Dataset (CK+)}

\begin{table}[t!]
\caption{Recognition accuracy on the Extended Cohn-Kanade (CK+) Dataset - 8 classes - A: Data Augmentation, D: Dropout}
\begin{center}
    \begin{tabular}{ | l | c  |}
    \hline
    \textbf{Method} & \textbf{Accuracy} \\ \hline    
    AURF \cite{liu2013aware} & 92.22\% \\ \hline
    AUDN \cite{liu2015inspired} & 93.70\%\\ 
    \hline \hline
    Zero-bias CNN & 78.2\% $\pm$ 5.7\%\\ \hline
    Zero-bias CNN+D & 82.3\% $\pm$ 4.0\%\\ \hline
    Zero-bias CNN+A & 94.6\% $\pm$ 3.3\% \\ \hline
    \textbf{Zero-bias CNN+AD} & \textbf{95.1\% $\pm$ 3.1\%}\\  
    \hline
    \end{tabular}
\end{center}
\label{tab:eight_class}
\end{table}

\begin{table}[t!]
\caption{Recognition accuracy on the Extended Cohn-Kanade (CK+) Dataset - 6 classes - A: Data Augmentation, D: Dropout}
\begin{center}
    \begin{tabular}{ | l | c  |}
    \hline
    \textbf{Method} & \textbf{Accuracy} \\ \hline
    CSPL \cite{zhong2012learning} & 89.89\% \\ \hline
    LBPSVM \cite{shan2009facial} & 95.10\% \\ \hline 
    \textbf{BDBN \cite{liu2014facial}} & \textbf{96.70\%} \\ 
    \hline \hline
    Zero-bias CNN+AD & 95.7\% $\pm$ 2.5\%\\  
    \hline
    \end{tabular}
\end{center}
\label{tab:six_class}
\end{table}

We now present our results on the CK+ dataset. The CK+ dataset usually contains eight labels  (anger, contempt, disgust, fear, happy, neutral, sad, and surprise). However, many works \cite{zhong2012learning, shan2009facial, liu2014facial} ignore the samples labeled as neutral or contempt, and only evaluate on the six basic emotions. Therefore, to ensure fair comparison, we trained two separate models. We present the eight class model results in Table \ref{tab:eight_class} and the six class model results in Table \ref{tab:six_class}. For the eight class model, we conduct the same study we did on the TFD and we observe rather similar results. Once again, regularization appears to play a significant role in obtaining good performance. Data augmentation gives a significant boost in performance ($16.4\%$) and when combined with dropout, leads to a $16.9\%$ increase. For the eight class and six class models, we achieve state-of-the-art and near state-of-the-art accuracy respectively on the CK+ dataset.

%

\subsection{Visualization of higher-level neurons}
Now, with a strong discriminative model in hand, we will analyze which facial regions the neural network identifies as the most discriminative when performing classification. To do this, we employ the visualization technique presented by Zeiler and Fergus in \cite{zeiler2014visualizing}. 

For each dataset, we consider the third convolutional layer and for each filter, we find the N images in the chosen split's training set that generated the strongest magnitude response. We then leave the strongest neuron high and set all other activations to zero and use the deconvolutional network to reconstruct the region in pixel space. For our experiments, we chose N=10 training images. 

We further refine our reconstructions by employing a technique called "Guided Backpropagation" proposed by Springenberg et al. in \cite{springenberg2014striving}. "Guided Backpropogation" aims to improve the reconstructed spatial patterns by not solely relying on the masked activations given by the top-level signal during deconvolution but by also incorporating knowledge of which activations were suppressed during the forward pass. Therefore, each layer's output during the deconvolution stage is masked twice: (i) once by the ReLU of the deconvotional layer and (ii) again by the mask generated by the ReLU of the layer's matching convolutional layer in the forward pass.

\begin{figure*}[t!]
  \centering
  \centerline{\includegraphics[width=17cm, height=17cm]{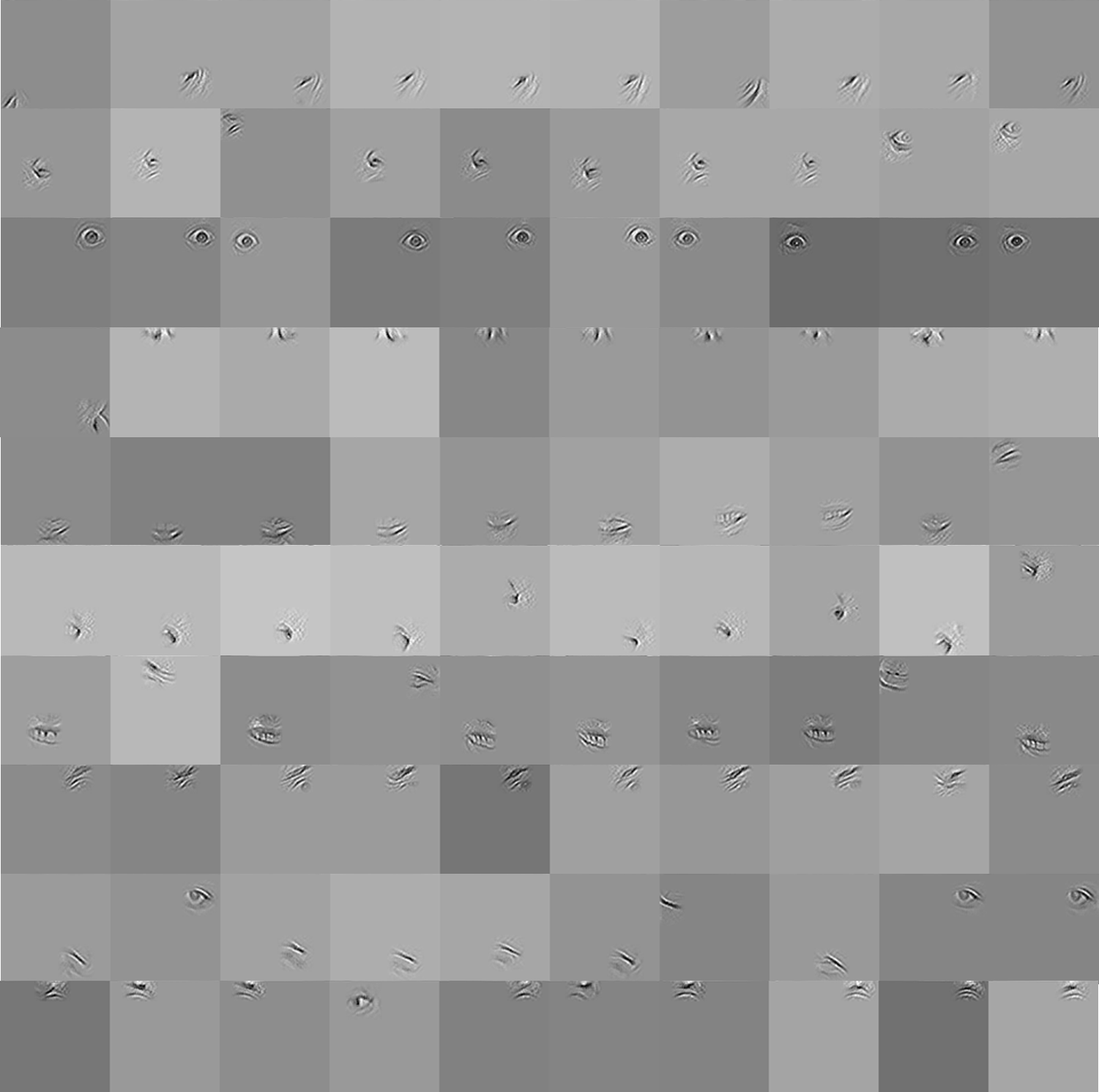}}
  \vspace{0.2cm}
  \caption{Visualization of spatial patterns that activate 10 selected filters in the conv3 layer of our network trained on the Toronto Face Dataset (TFD). Each row corresponds to one filter in the conv3 layer. We display the top 10 images that elicited the maximum magnitude response. Notice that the spatial patterns appear to correspond with some of the Facial Action Units.} 
\label{fig:zeiler_plots_tfd}
\end{figure*}

First, we will analyze patterns discovered in the Toronto Face Dataset (TFD). In Figure \ref{fig:zeiler_plots_tfd}, we select 10 of the 256 filters in the third convolutional layer and for each filter, we present the spatial patterns of the top-10 images in the training set. From these images, the reader can see that several of the filters appear to be sensitive to regions that align with several of the Facial Actions Units such as: AU12: Lip Corner Puller (row 1), AU4: Brow Lowerer (row 4), and AU15: Lip Corner Depressor (row 9). 

\begin{figure*}[t!]
  \centering
  \centerline{\includegraphics[width=17cm, height=17cm]{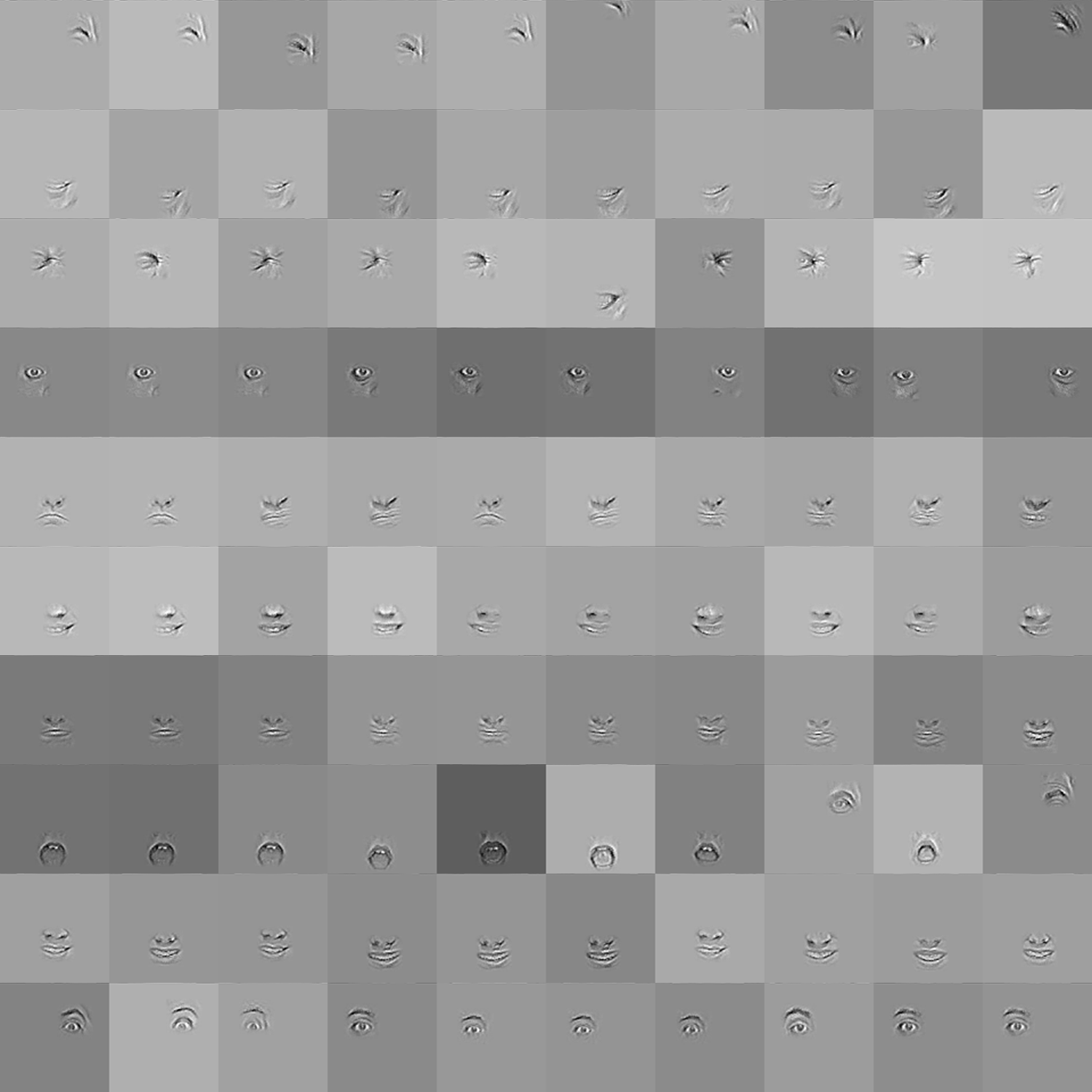}}
  \vspace{0.2cm}
  \caption{Visualization of spatial patterns that activate 10 selected filters in the conv3 layer of our network trained on the Cohn-Kanade (CK+) dataset. Each row corresponds to one filter in the conv3 layer. Once again, we display the top 10 images that elicited the maximum magnitude response. Notice that the spatial patterns appear to have very clear correspondences with some of the Facial Action Units.} 
\label{fig:zeiler_plots_ck_plus}
\end{figure*}

Next, we display the patterns discovered in the CK+ dataset. In Figure \ref{fig:zeiler_plots_ck_plus}, we, once again, select 10 of the 256 filters in the third convolutional layer and for each filter, we present the spatial patterns of the top-10 images in the training set. The reader will notice that the CK+ discriminative spatial patterns are very clearly defined and correspond nicely with Facial Action Units such as: AU12: Lip Corner Puller (rows 2, 6, and 9), AU9: Nose Wrinkler (row 3) and AU27: Mouth Stretch (row 8).  
 

\begin{table}
\caption{Correspondences between CK+ visualization plots shown in Figure \ref{fig:zeiler_plots_ck_plus} and the FAU whose activation distribution had the highest KL divergence value. The KL divergence values of all the FAUs computed for each filter are shown in Figure \ref{fig:fau_kl_div_bar_charts}.}
\begin{center}
    \begin{tabular}{| c | c |}
    \hline
    \makecell{\bf Filter \\ \bf Number } & \makecell{ \bf FAU with the Largest \\ \bf KL Divergence Value}  \\ \hline
    1 & AU25: Lips Part \\ \hline    
    2 & AU12: Lip Corner Puller \\ \hline
    3 & AU9: Nose Wrinkler \\ \hline
    4 & AU5: Upper Lid Raiser \\ \hline    
    5 & AU17: Chin Raiser \\ \hline
    6 & AU12: Lip Corner Puller \\ \hline
    7 & AU24: Lip Pressor \\ \hline
    8 & AU27: Mouth Stretch \\  \hline
    9 & AU12: Lip Corner Puller \\ \hline
    10 & AU1: Inner Brow Raiser \\  \hline        	
    \end{tabular}
\end{center}
\label{tab:ck_plus_AU_correspondence_list}
\end{table}

\subsection{Finding Correspondences Between Filter Activations and the Ground Truth Facial Action Units (FAUs)}

In addition to categorical labels (anger, disgust, etc.), the CK+ dataset also contains labels that denote which FAUs are present in each image sequence. Using these labels, we now present a preliminary experiment to verify that the filter activations/spatial patterns learned by the CNN indeed match with the actual FAUs shown by the subject in the image. Our experiment aims to answer the following question: For a particular filter i, which FAU j has samples whose activation values most strongly differ from the activations of samples that do not contain FAU j, and does that FAU accurately correspond with the visual spatial patterns that maximally excite filter i?

Given a training set of M images ($X$) and their corresponding FAU labels ($Y$), let $ F_{\ell i}(x)$ be the activations of sample x at layer $\ell$ for filter $i$. Since we are examining the 3rd convolutional layer in the network, we set $\ell=3$. Then, for each of the 10 filters visualized in Figure \ref{fig:zeiler_plots_ck_plus}, we do the following: 
\begin{enumerate}
\item[(i)] We consider a particular FAU j and place the samples $X$ that contain j in set S where: \\
$ S = \{ x_m \: | \: j \in y_m\}, \: \forall m \in \{1, ..., M\} $
\item[(ii)] We then build a histogram of the maximum activations of the samples that contained FAU j: $ \\ Q_{ij}(x) = P(F_{3i}(x) \: | \: S), \: \forall (x, y) \in (X, Y)$ 
\item[(iii)] We then, similarly, build a distribution over maximum activations of the samples that do not contain FAU j: \\$ R_{ij}(x) = P(F_{3i}(x) \: | \: S ^{c}), \: \forall (x, y) \in (X, Y) $
\item[(iv)] We compute the KL divergence between $Q_{ij}(x)$ and $R_{ij}(x)$, $ D_{KL}(Q_{ij} \: \| \: R_{ij})$, and repeat the process for all of the other FAUs.

\end{enumerate}

\begin{figure*}[t!]
\centering
 \vspace{-0.5cm}
\begin{minipage}[b]{0.48\linewidth}
  \centering
  \centerline{\includegraphics[trim = 0mm -5mm 20mm 190mm, clip, width=10cm, height=4.1cm, keepaspectratio] {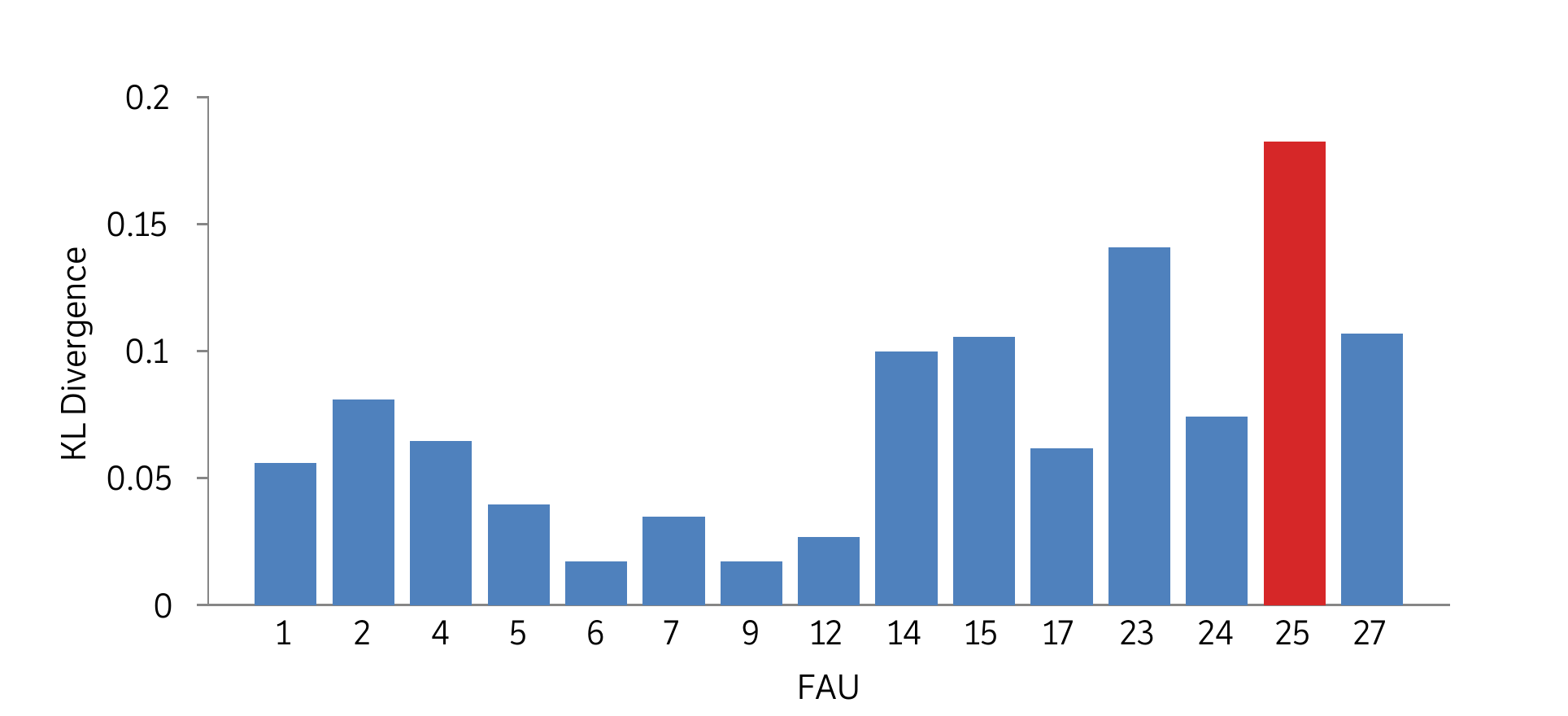}}
  \vspace{-.25cm}
  \centerline{Filter 1}
\end{minipage}
\hfill
\begin{minipage}[b]{0.48\linewidth}
  \centering
  \centerline{\includegraphics[trim = 0mm -5mm 20mm 190mm, clip, width=10cm, height=4.1cm, keepaspectratio]{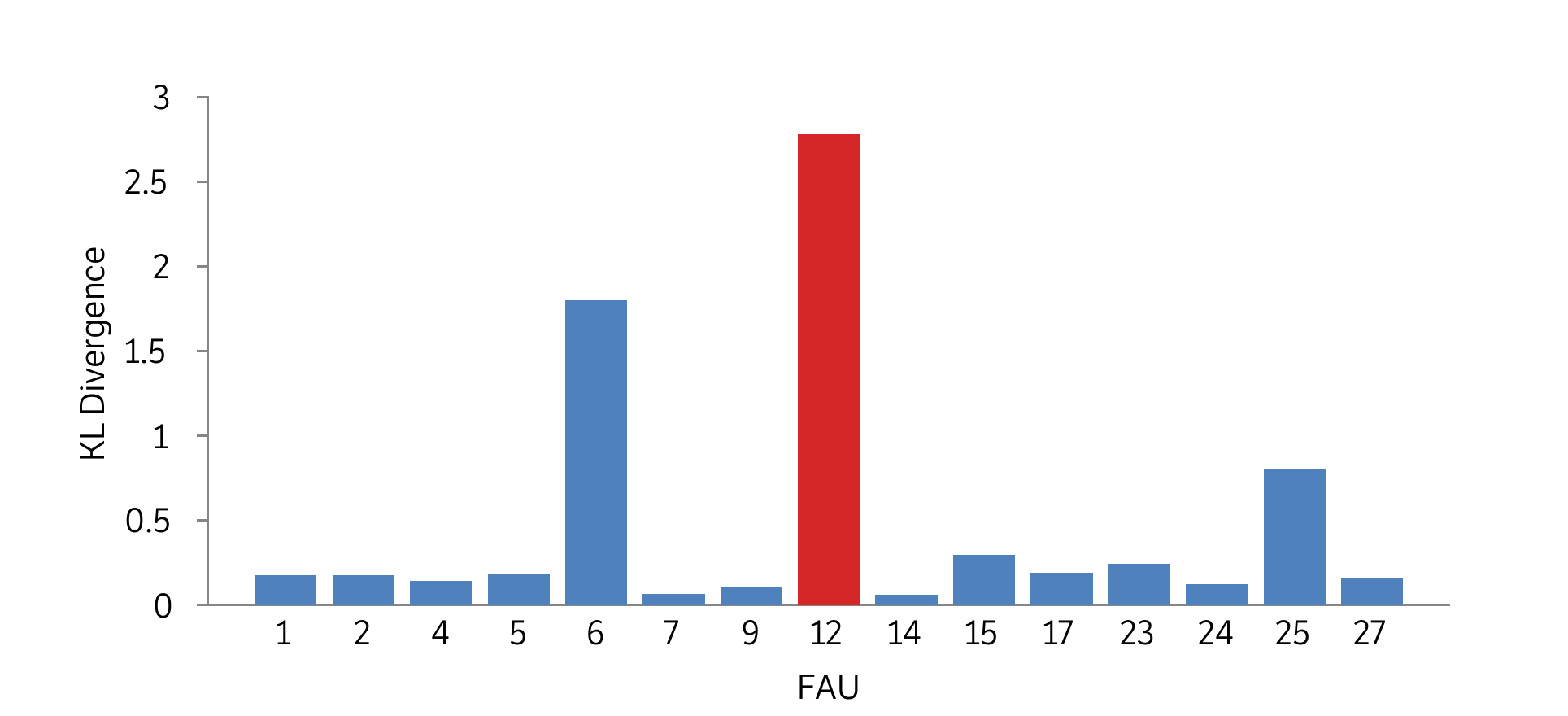}}
  \vspace{-.25cm}
  \centerline{Filter 2}
\end{minipage}

\vspace{-.25cm}

\begin{minipage}[b]{0.48\linewidth}
  \centering
  \centerline{\includegraphics[trim = 0mm -5mm 20mm 190mm, clip, width=10cm, height=4.1cm, keepaspectratio]{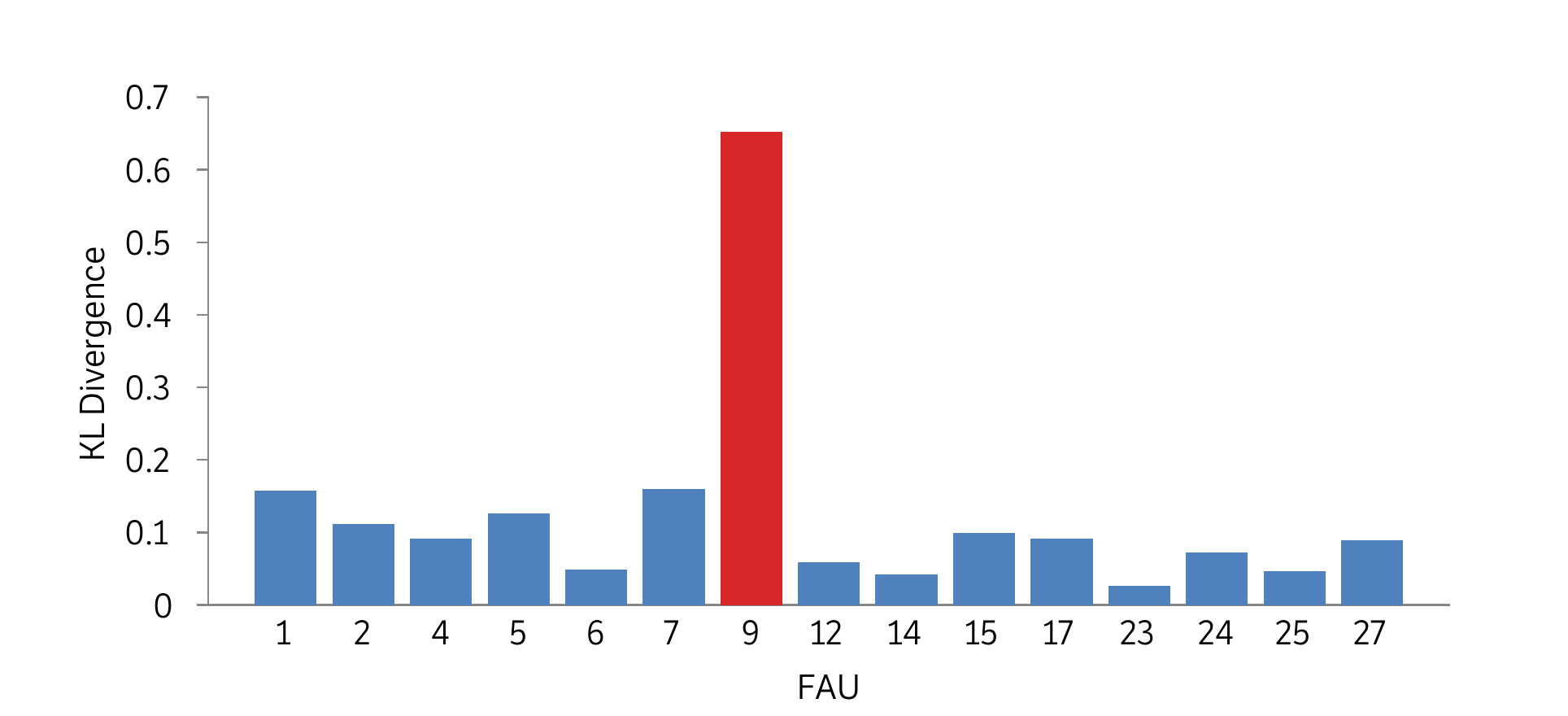}}
  \vspace{-.25cm}
  \centerline{Filter 3}
\end{minipage}
\hfill
\begin{minipage}[b]{0.48\linewidth}
  \centering
  \centerline{\includegraphics[trim = 0mm -5mm 20mm 190mm, clip, width=10cm, height=4.1cm, keepaspectratio]{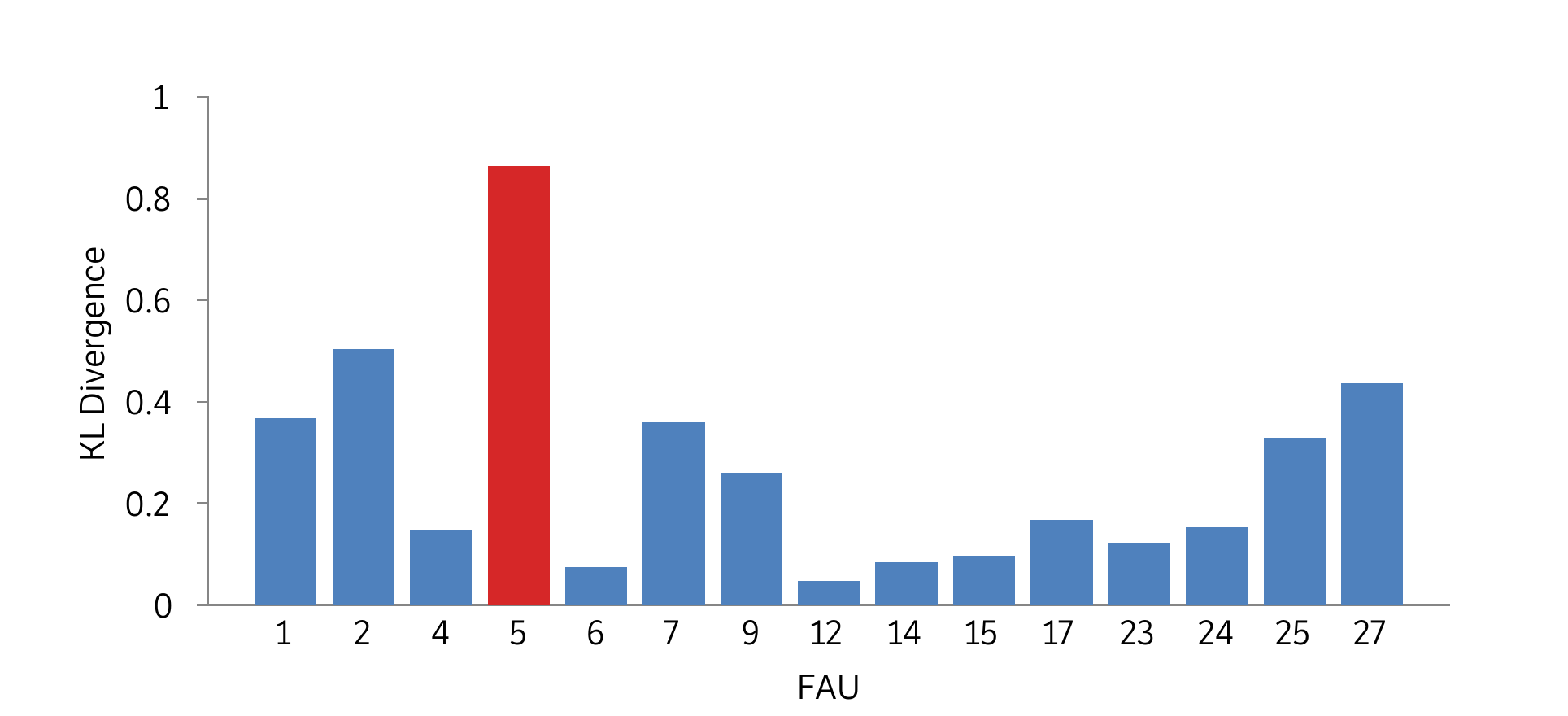}}
  \vspace{-.25cm}
  \centerline{Filter 4}
\end{minipage}

\vspace{-.25cm}

\begin{minipage}[b]{0.48\linewidth}
  \centering
  \centerline{\includegraphics[trim = 0mm -5mm 20mm 190mm, clip, width=10cm, height=4.1cm, keepaspectratio] {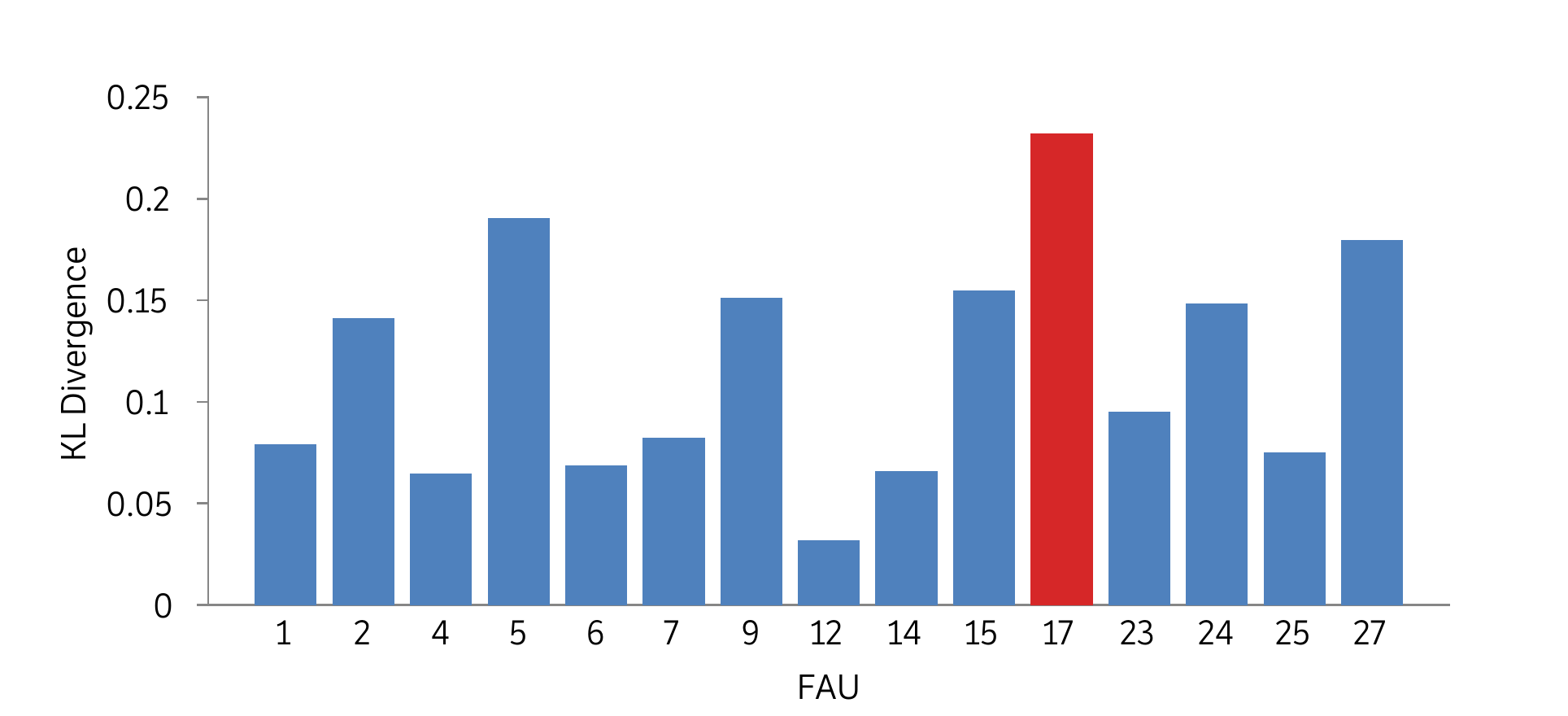}}
  \vspace{-.25cm}
  \centerline{Filter 5}
\end{minipage}
\hfill
\begin{minipage}[b]{0.48\linewidth}
  \centering
  \centerline{\includegraphics[trim = 0mm -5mm 20mm 190mm, clip, width=10cm, height=4.1cm, keepaspectratio]{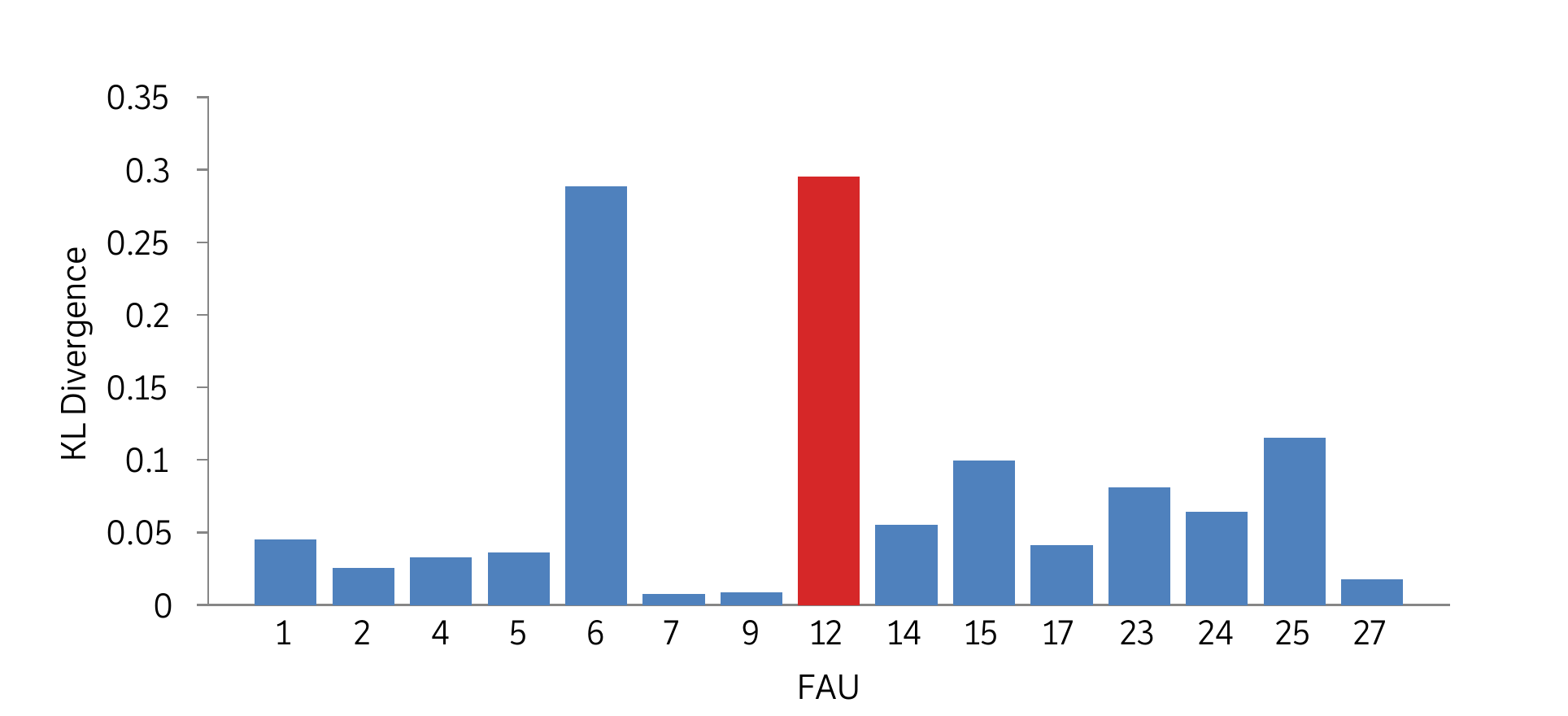}}
  \vspace{-.25cm}
  \centerline{Filter 6}
\end{minipage}

\vspace{-.25cm}

\begin{minipage}[b]{0.48\linewidth}
  \centering
  \centerline{\includegraphics[trim = 0mm -5mm 20mm 190mm, clip, width=10cm, height=4.1cm, keepaspectratio]{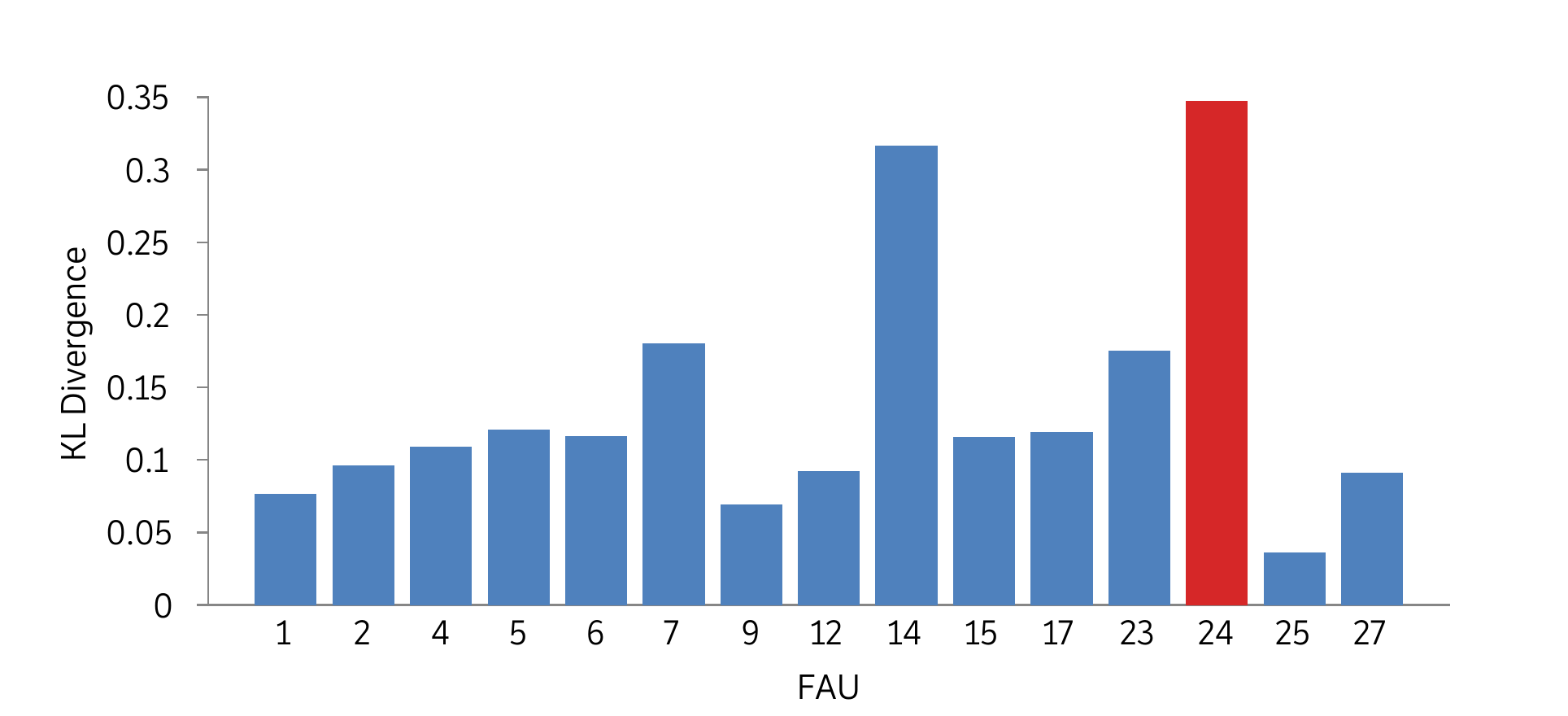}}
  \vspace{-.25cm}
  \centerline{Filter 7}
\end{minipage}
\hfill
\begin{minipage}[b]{0.48\linewidth}
  \centering
  \centerline{\includegraphics[trim = 0mm -5mm 20mm 190mm, clip, width=10cm, height=4.1cm, keepaspectratio]{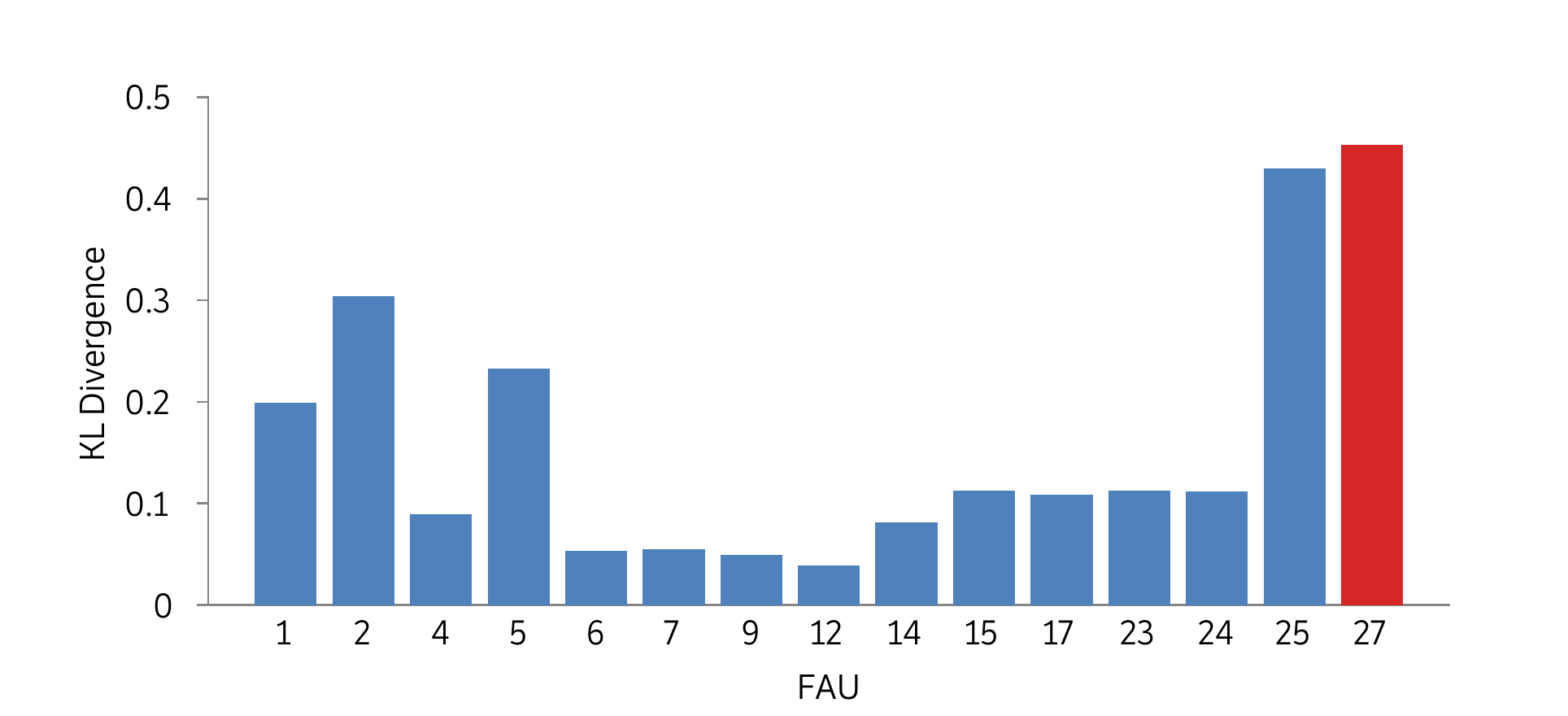}}
  \vspace{-.25cm}
  \centerline{Filter 8}
\end{minipage}

\vspace{-.25cm}

\begin{minipage}[b]{0.48\linewidth}
  \centering
  \centerline{\includegraphics[trim = 0mm -5mm 20mm 190mm, clip, width=10cm, height=4.1cm, keepaspectratio]{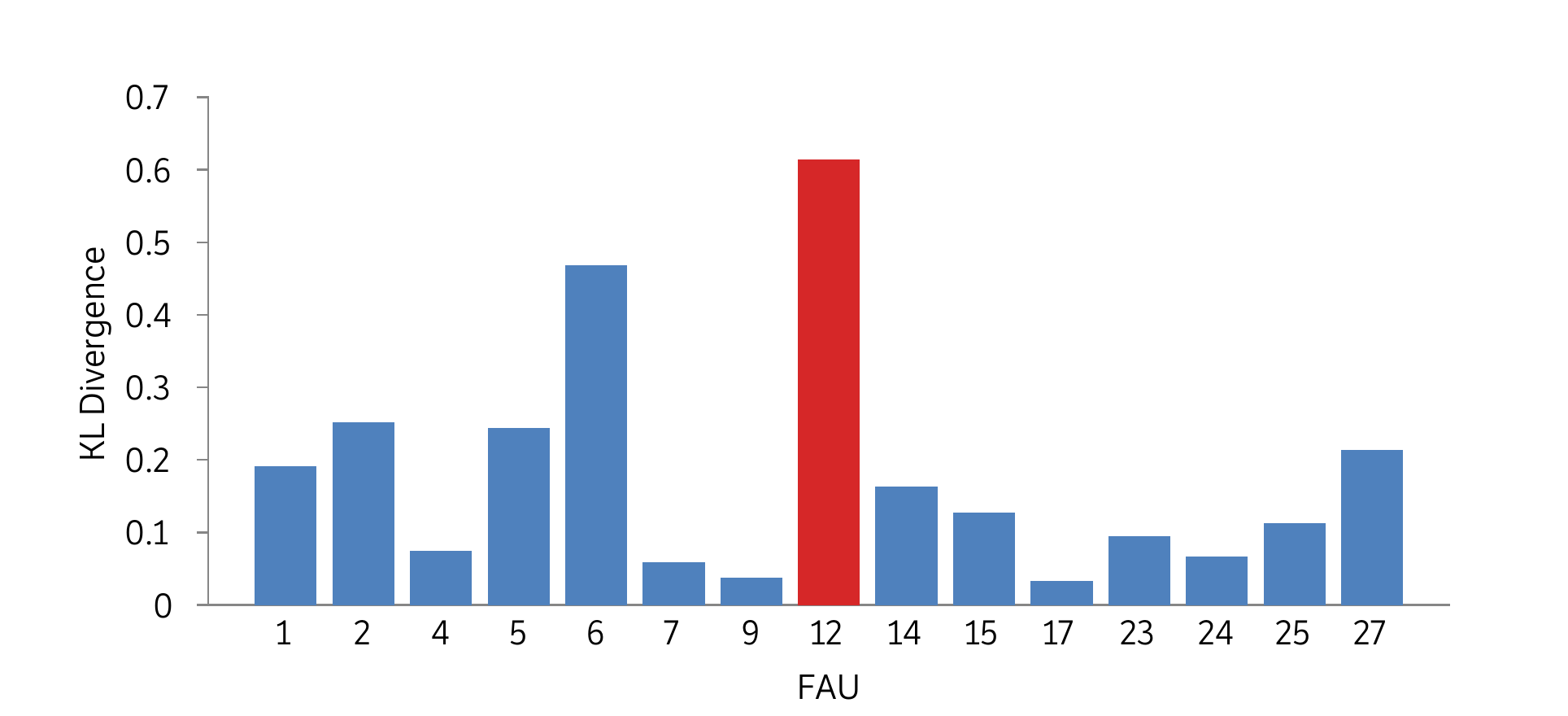}}
  \vspace{-.25cm}
  \centerline{Filter 9}
\end{minipage}
\hfill
\begin{minipage}[b]{0.48\linewidth}
  \centering
  \centerline{\includegraphics[trim = 0mm -5mm 20mm 190mm, clip, width=10cm, height=4.1cm, keepaspectratio]{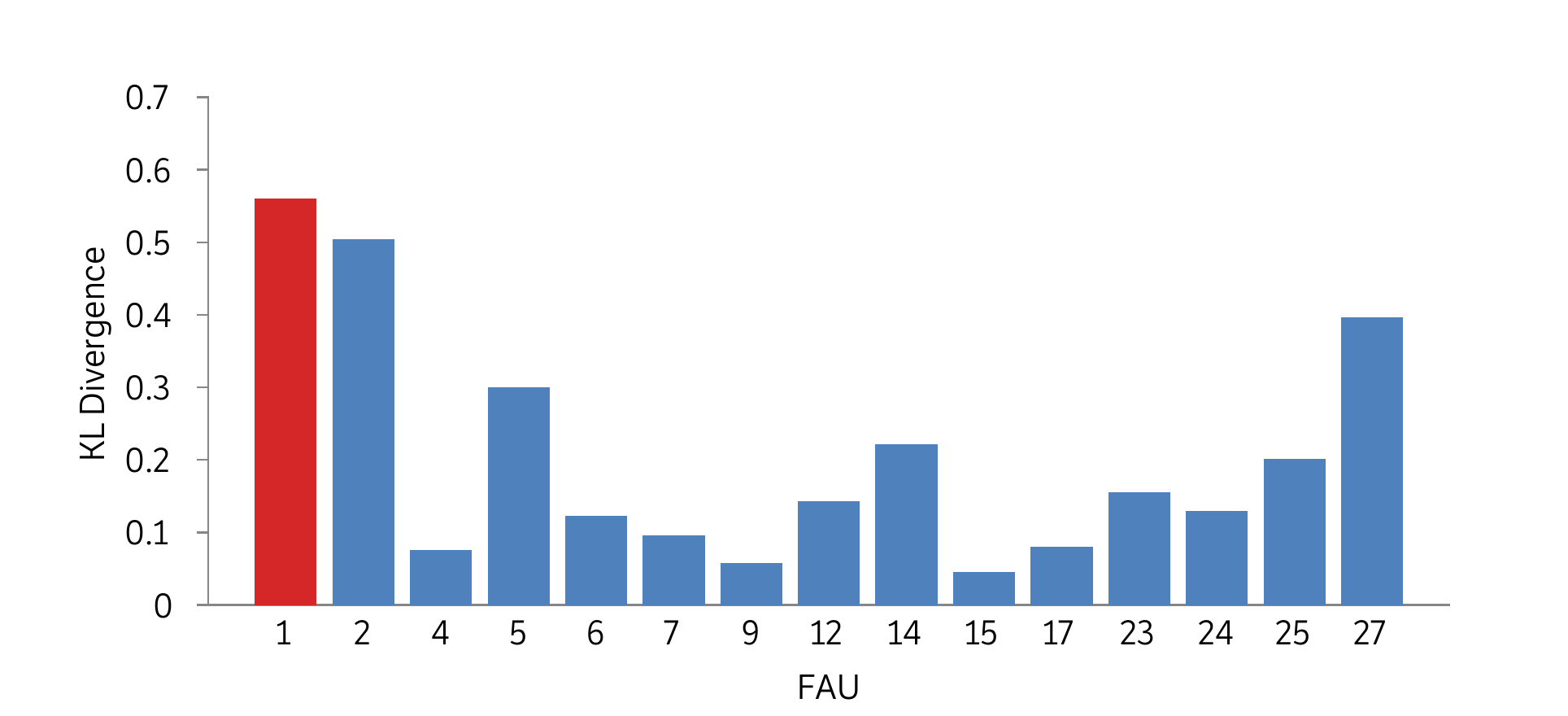}}
  \vspace{-.25cm}
  \centerline{Filter 10}
\end{minipage}

\vspace{0.5cm}
\caption{Bar charts showing which FAUs lead to the strongest shifts in the activation distributions of particular filters in the CNN. For each of the 10 filters visualized in Figure \ref{fig:zeiler_plots_ck_plus}, we build histograms over the activations of training samples that contain a specific FAU j, and the activations of samples that do not contain FAU j. We then compute the KL divergence between the two distributions and plot them for each FAU above. The FAU with the largest KL divergence is displayed in red and its corresponding name is given in Table \ref{tab:ck_plus_AU_correspondence_list}. (Best viewed in color). }
\label{fig:fau_kl_div_bar_charts}
\end{figure*}

Figure \ref{fig:fau_kl_div_bar_charts} shows the bar charts of the KL divergences computed for all of the FAUs for each of the 10 filters displayed in Figure \ref{fig:zeiler_plots_ck_plus}. The FAU with the largest KL divergence value is denoted in red and its corresponding name is documented in Table \ref{tab:ck_plus_AU_correspondence_list} for each filter. From these results, we see that in the majority of the cases, the FAUs listed in Table \ref{tab:ck_plus_AU_correspondence_list} match the facial regions visualized in Figure \ref{fig:zeiler_plots_ck_plus}. This means that the samples that appear to strongly influence the activations of these particular filters are indeed those that possess the AU shown in the corresponding filter visualizations. Thus, we show that certain neurons in the neural network implicitly learn to detect specific FAUs in face images when given a relatively "loose" supervisory signal (i.e. emotion type: anger, happy, sad, etc.). 

What is most encouraging is that these results appear to confirm our intuitions about how CNNs work as appearance-based classifiers. For instance, filter 2, 6, and 9 appear to be very sensitive to patterns that correspond to AU 12. This is not surprising as AU 12 (Lip Corner Puller) is almost always associated with smiles and from the visualizations in Figure \ref{fig:zeiler_plots_ck_plus}, a subject often shows their teeth when smiling, a highly distinctive appearance cue. Similarly, for filter 8, it is not surprising that FAU 25 (Lips Part) and FAU 27 (Mouth Stretch) had the most different activation distributions given that the filter's spatial patterns corresponded to the "O" shape made by the mouth region in surprised faces, another visually salient cue.

%% file: Sections/conclusions.tex

In this work, we showed both qualitatively and quantitatively that CNNs trained to do emotion recognition are indeed able to model high-level features that strongly correspond to FAUs. Qualitatively, we showed which portions of the face yielded the most discriminative information by visualizing the spatial patterns that maximally excited different filters in the convolutional layers of our learned networks. Meanwhile, quantitatively, we correlated the numerical activations of the visualized filters with the subject's actual facial movements using the FAU labels given in the CK+ dataset. Finally, we demonstrated how a zero-bias CNN can achieve state-of-the-art recognition accuracy on the extended Cohn-Kanade (CK+) dataset and the Toronto Face Dataset (TFD).